\documentclass{article}

\usepackage[final]{template_custom}
\usepackage[utf8]{inputenc} 
\usepackage[T1]{fontenc}  
\usepackage{url}     
\usepackage{booktabs}    
\usepackage{amsfonts}       
\usepackage{nicefrac}    
\usepackage{microtype}     
\usepackage{xcolor}         
\usepackage{subcaption} 
\usepackage{multirow}
\usepackage{array}

\usepackage{nicematrix}
\usepackage{tabularx}
\usepackage{colortbl}
\usepackage{makecell}

\usepackage{wrapfig}
\definecolor{citecolor}{HTML}{0071BC}
\definecolor{linkcolor}{HTML}{ED1C24}
\usepackage[pagebackref=false, breaklinks=true,letterpaper=true, colorlinks,citecolor=citecolor, linkcolor=linkcolor, bookmarks=false]{hyperref}
\usepackage{xspace}

\definecolor{background}{rgb}{1,1,1} 
\definecolor{softblue}{rgb}{0.8, 0.83, 1}
\definecolor{LightCyan}{rgb}{0.92,1,1}

\definecolor{softlowcost}{rgb}{0.2, 0.9, 0.4} 
\definecolor{softmediumcost}{rgb}{0.73, 0.73, 0.46} 
\definecolor{softhighcost}{rgb}{1.0, 0.2, 0.2} 

\newcolumntype{x}[1]{>{\centering\arraybackslash}p{#1pt}}
\newcolumntype{y}[1]{>{\raggedright\arraybackslash}p{#1pt}}
\newcolumntype{z}[1]{>{\raggedleft\arraybackslash}p{#1pt}}

\definecolor{darkergreen}{RGB}{21, 152, 56}
\definecolor{red2}{RGB}{252, 54, 65}

\newcommand\greenp[1]{\textcolor{darkergreen}{(#1)}}

\newcommand{\ModelName}{VisInContext\xspace}
\newcommand{\ModelFullName}{Visualized In-Context Text Processing\xspace}

\title{ Leveraging Visual Tokens for Extended Text Contexts in Multi-Modal Learning}

\author{Alex Jinpeng Wang$^{1}$ \quad Linjie Li$^{2}$ \quad Yiqi Lin$^{1}$ \quad Min Li$^{3}$ \\ \quad \bf Lijuan Wang$^{2}$ \quad \bf Mike Zheng Shou$^{1}$\\[3pt]
$^1$Show Lab, National University of Singapore \quad $^2$Microsoft Gen AI \quad $^3$Central South University \\
\url{http://fingerrec.github.io/visincontext}}

\begin{document}

\maketitle

\begin{abstract}
Training models with longer in-context lengths is a significant challenge for multimodal model due to substantial GPU memory and computational costs. 
This exploratory study does not present state-of-the-art models; rather, it introduces an innovative method designed to increase in-context text length in multi-modality large language models (MLLMs) efficiently. 
We present \ModelFullName(\ModelName), which processes long in-context text using visual tokens. 
This technique significantly reduces GPU memory usage and floating point operations (FLOPs) for both training and inferenceing stage. 
For instance, our method expands the pre-training in-context text length from 256 to 2048 tokens with nearly same FLOPs for a 56 billion parameter MOE model. 
Experimental results demonstrate that model trained with \ModelName delivers superior performance on common downstream benchmarks for in-context few-shot evaluation. 
Additionally, \ModelName~ is complementary to existing methods for increasing in-context text length and enhances document understanding capabilities, showing great potential in document QA tasks and sequential document retrieval.
\end{abstract}
\section{Introduction}

\begin{figure}[ht]
    \centering
    \begin{subfigure}[b]{0.48\textwidth}
        \centering  \includegraphics[width=\linewidth]{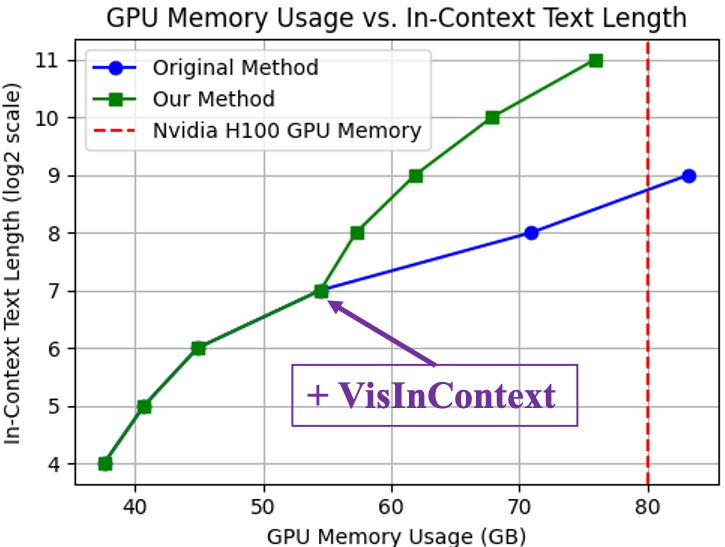} 
        \caption{
        GPU memory consumption.}
        \label{fig:sub1}
    \end{subfigure}
    \hfill 
    \begin{subfigure}[b]{0.48\textwidth}
        \centering
\includegraphics[width=\linewidth]{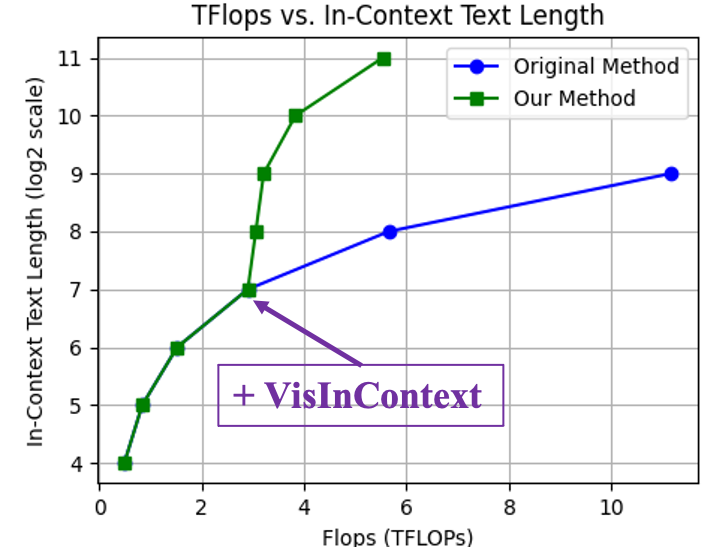}
        \caption{
        The flops comparison.}
        \label{fig:sub2}
    \end{subfigure}
    \caption{
    \textbf{
    \small 
    \ModelName~significantly increases the in-context text length from 256 to 2048 during pre-training
    on NVIDIA H100 GPU.} 
    For our method, we incorporate \ModelName~after 128 text tokens. 
    We implement PyTorch Flamingo~\cite{flamingo} models with different in-context length during pre-training.
    The language model is a 56B MOE~\cite{moe} model loaded with 4-bit quantization and the batch size on each GPU is 32 with FP16.
    We train the model with DeepSpeed~\cite{deepspeed} Zero-2.
    }
    \label{fig:motivation}
\end{figure}

Large Language Models (LLMs), such as OPT, Mistral, and LLaMA-2~\cite{opt,mistral,llama2}, have significantly advanced the field of Natural Language Processing (NLP). 
These advancements are partly due to the increased capability of LLMs to process long contexts, from 512 tokens~\cite{bert} up to 16K tokens~\cite{llama2}. 
Building on these developments, recent multi-modal learning research~\cite{flamingo,adept_fuyu8b,openflamingo,deepseek} has shifted focus from simple image-text pairs, like those in CC3M~\cite{cc3m} and LAION-400M~\cite{laion400m}, to more complex and lengthy interleaved document datasets. 
Examples include web corpora like MMC4~\cite{mmc4} and the OBELICS~\cite{obelics} dataset, as well as PDF corpora like DocVQA~\cite{docvqa}.

However, training models on these complex datasets presents significant challenges due to the increased GPU memory and computational demands of extended contexts. 
For instance, while processing just 5M data items from MMC4 and 10M from the OBELICS dataset, OpenFlamingo-9B~\cite{openflamingo} resorted to sub-sampling text and processing only 256 tokens at a time, yet it still requires 32 80GB A100 GPUs for over three days. 
This highlights the need for more computation-efficient methods to handle long context lengths effectively.

In the domain of LLMs, two popular methods to extend context length are the use of memorizing banks~\cite{memorizingtransformer} and novel self-attention mechanisms~\cite{flashattention, ring}. 
These methods have inspired advancements in the multi-modality domain as well. 
For example, the Large World Model~\cite{lvm} introduces Ring Attention~\cite{ring}, and MA-LMM~\cite{malmm} employs memory banks to process long video understanding tasks. 
While these techniques have shown promise, our approach aims to increase in-context text length by leveraging the strengths of visual encoders in MLLMs.
\textbf{We first observe that existing MLLMs usually exploit a much lighter visual encoders, compared to its text decoders}. 
For instance, Flamingo-9B consists of a 304.4M ViT-L/16~\cite{threethings} as image encoder, and a 7.1B  Chinchilla~\cite{flamingo} model as the text decoder. 
Additionally, previous works~\cite{clip,parrot} have demonstrated that visual encoders trained on paired image-text data also \textbf{exhibit emergent OCR capabilities}.

Motivated by these observations, we propose
\ModelFullName (\ModelName), a method that uses visual tokens to process extended textual contexts, which is complementary of existing methods in extending context length.
Specifically, we \textbf{convert long textual content into images and use the visual encoders to extract textual representations}. 
In this way, we can efficiently and effectively enable models with much longer text contexts, as shown in Figure~\ref{fig:motivation}.
With \ModelName, we show that the in-context text length can be increased  by 7 times over the competing baseline. 
Additionally, we observe almost the same overall computation FLOPs even as in-context length extends significantly. Our extensive experiments will also show that  \ModelName renders superior model performance on conventional in-context few-shot evaluations and document understanding, with much lower computational cost.

\textbf{Contributions.} In summary, our contributions are as follows: 
\emph{i.} We introduce \ModelFullName (\ModelName), a novel method that increases in-context text length using visual tokens.
\ModelName directly compresses text context at input-level, which is complementary to existing 
 techniques with improved self-attention or memory banks.
\emph{ii.} We demonstrate that \ModelName~is effective for both training and inference stage with much lower computational cost.
\emph{iii.} With extended text context brought by \ModelName, our model improves the average in-context few-shot performance from 55.8\% to 57.8\% over the competing baseline.
\emph{iv.} As a byproduct, our method also shows great potential in document understanding on popular document QA tasks and our newly proposed sequential document retrieval task.

\section{Method}

The goal of \ModelName is to process in-context text using visual tokens so that the model can handle long text context more efficiently. 
We primarily base our study on Flamingo-based architecture~\cite{flamingo,openflamingo,obelics}, as it has shown success in improving a model's ability to learn from long multimodal context that contains arbitrarily interleaved text and images.

\subsection{Terminology} 
Before diving into model details, we define the following terms:

\textit{In-context Text Length}: The \textbf{actual length of text tokens observed by the model within a document}.

\textit{Text Token Length}: The length of the \textbf{text sequence input directly to the LLM}, corresponding to the token count of this sequence.

With \ModelName, the \textit{In-context Text Length} is greater than the text token length, as part of the text is represented using visual tokens.

\subsection{Overall Architecture}

\begin{figure}
    \centering
\includegraphics[width=\linewidth]{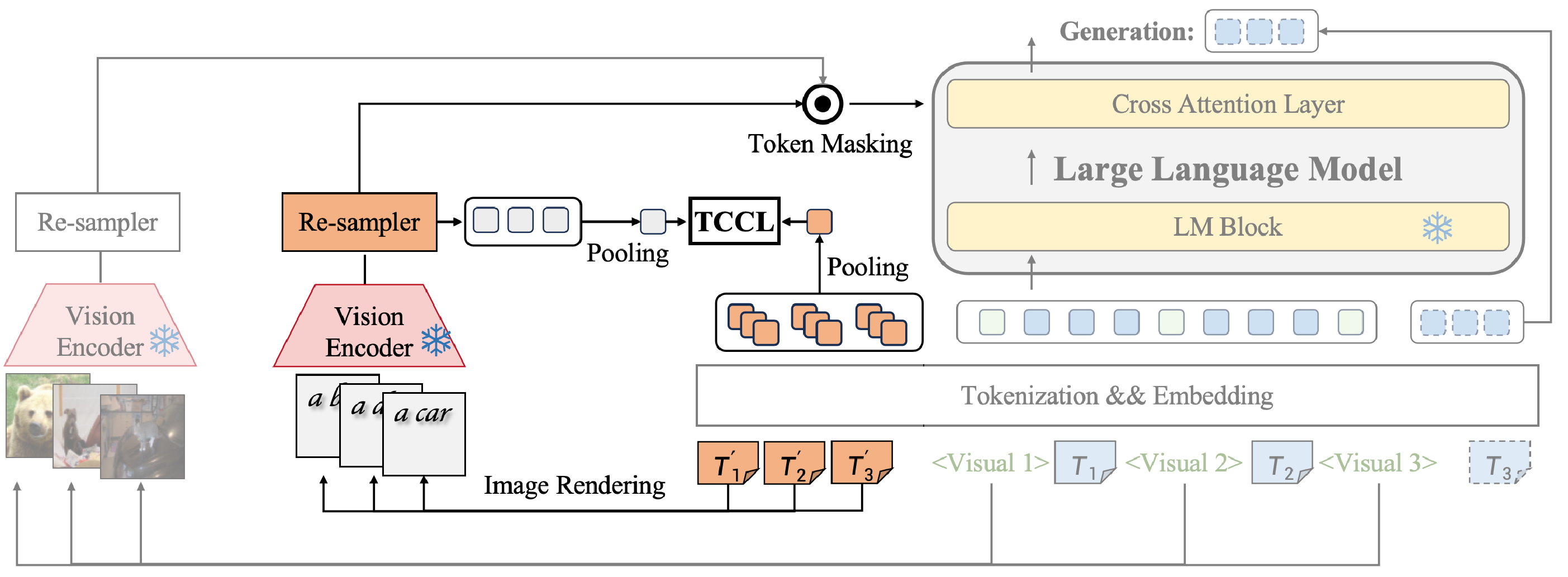}
    \caption{
    \small 
    \textbf{\ModelName Pipeline}.
    The \ModelName pipeline builds upon the Flamingo model for in-context few-shot modeling (represented in gray).
    \ModelName processes interleaved image-text data by rendering portions of the in-context text into images. This approach maintains the \textit{Text Token Length} of the model while allowing for a significantly extended \textit{In-context Text Length}.
    }
    \vspace{-1em}
    \label{fig:overall}
\end{figure}

The implementation and architecture of \ModelName are shown in Figure~\ref{fig:overall}.
It is based on a dual-stream encoder model that integrates both visual and textual data. 
To effectively handle long interleaved data, we use a pre-sampling strategy as in Flamingo-style works~\cite{flamingo,openflamingo,obelics}.
Specifically, we sample $m$ images, denoted as $I_1, I_2, \ldots, I_m \in I$, along with their corresponding texts $T_1, T_2, \ldots, T_m \in T$. 
These tokens are concatenated, resulting in a sequence of about 256 tokens.
However, since the overall length of a web document is generally much longer than 256 tokens (\textit{In-context Text Length $\geq$ Text Token Length}), this sampling approach can lead to the omission of a lot of related text context.

To address this issue, we convert these omitted text context into visual signals by rendering them into images.
We first concatenate all omitted text segments and divide them into $M$ parts to render text images, named $T_{1}^{'}, T_{2}^{'}, \ldots, T_{m}^{'} \in T'$.
Both the original images and the text-rendered images are then processed through a shared frozen vision encoder.
Then, we employ two learnable resamplers to extract a fixed number of tokens from both the raw and text-rendered image features, respectively.
To facilitate the model to learn from rendered text images, we introduce two novel model designs,  Token Masking mechanism and Text-Centric Contrastive Learning (TCCL).
Token Masking allows the model to only read from text image tokens by masking the raw image tokens with masking ratio 1, which ensures that the model won't simply be ignoring the text images during training, hence can learn the association between the rendered text images $\{T_{i}^{'}\}$ and the text tokens $\{T_{i}\}$. 
TCCL aligns the visual text representation from the resampler with the embeddings extracted from text tokenizers in LLM, which reduces the gap between our visual text tokens and the text tokens the LLM is trained to perceive.
With these designs, \ModelName not only reduces computational demands—as evidenced by a reduction in flops and inference time—but also improves the OCR ability, as we will show in our experiments.

\subsection{Text Rendering}
This module converts textual data into a visually rich RGB format, specifically rendering the text into an image size of $p_h \times n p_w$, where $n$ is the number of patches. 
We employ the \textit{HERSHEY} font at a size of 10px. 
On average, one 16x16 patch accommodates approximately 1.5 OPT text tokens. A 224x224 text image  contains about 294 text tokens.
Consequently, a visual encoder 
operating on this rendered text image 
requires only $1/3$ of tokens to encode an equivalent amount of text, compared to the text tokenizer in language models.
The vision encoder is quite lightweight ViT-L (340M) compared to language model MOE (56B), which makes \textbf{the processing of rendered text images significantly more efficient than directly inputting the text into a language model}.

\subsection{Token Masking}
In our initial experiments, we find that combining tokens from raw images and text images directly led to the network disregarding the text-image input.
To address this issue, we introduce  a Token Masking strategy to force the model to learn text semantics from visual inputs.
During pretraining, the raw image and text image are first encoded into the same number of tokens after resampler, and then we mask the raw image tokens with a pre-defined probability. When masking out the raw image tokens, the model can focus on learning the association between rendered text images and the complementary text tokens.
At inference time, we add the text-image tokens and image tokens together, to allow the model effectively leverage information from both sources.

\subsection{Text-Centric Contrastive Loss (TCCL)}

\paragraph{Motivation.}
Given that the vision encoder, typically a frozen Vision Transformer (ViT)~\cite{vit}, never observes rendered text images during pretraining, it may struggle to derive text semantics from pixels. 
To mitigate this issue, we introduce a new training objective, Text-Centric Contrastive Loss (TCCL). This objective aims to guide 
the resampler on rendered text images to interpret visual representations of text with a proficiency comparable to traditional text tokenizers, so that  the textual semantics can be effective extracted from the rendered text images. 

\paragraph{Mechanism.}
TCCL utilizes raw text token embeddings from the text tokenizer as soft supervision signals to supervise the resampler to learn text-centric representation.
To reduce the global semantic gap between text image embeddings and text token embeddings, we first aggregate these embeddings with average pooling and then align them with TCCL.
Intuitively, TCCL is designed to turn the joint of the vision encoder and resampler into a ``visual" text tokenizer, as it promotes the text image embeddings to share a similar global semantic as the text token embeddings.
The core of TCCL is formulated as a contrastive loss:
\begin{equation}
\mathcal{L}_{ij} = -\log \left(\frac{\exp(\text{sim}(f_{v_i}, f_{t_j})/\tau)}{\sum_{k=1}^{N}\exp(\text{sim}(f_{v_i}, f_{t_k})/\tau)}\right)
\end{equation}
Where $\mathcal{L}_{ij}$ denotes the contrastive loss for comparing the $i^{th}$ text image against the $j^{th}$ text, $f_{v_i}$ and $f_{t_j}$ represent the feature embeddings of the $i^{th}$ text image and $j^{th}$ text, respectively. 
$\tau$ is a parameter that control the sharpness of the output distribution. Note that $f_{v_i}$ and $f_{t_i}$ are different features extracted from the same text, as the $i^{th}$ text image is a direct rendering of the $i^{th}$ text.
\section{Experiment}

\subsection{Experimental Setup}

\paragraph{Pretraining.} We validate \ModelName with Open-Flamingo~\cite{openflamingo} and CosMo~\cite{cosmo}. 
To enhance computational efficiency, all models utilize float16 precision. 
For the 56B MOE~\cite{moe} model, we employ DeepSpeed's~\cite{deepspeed} Zero-2 stage with CPU offloading and further optimize the model by quantizing it to 4-bit precision~\footnote{The implementation is from https://github.com/TimDettmers/bitsandbytes.}. 
We also use Flash Attention~\cite{flashattention} to further improve memory efficiency.
For all other experiments, we train the model using DeepSpeed Zero-2 without CPU off-loading.
The Open-Flamingo 9B baseline is based on Mistral7B~\cite{mistral}.

Our pretraining dataset includes a 180M subset of DataComp1B~\cite{datacomp}, MMC4~\cite{mmc4}, the OBELICS~\cite{obelics} dataset, and OCR Rendered Text~\cite{renderedtext2023}. (More details are provided in the Appendix) 
For each input document or image-text pair, we render a text sequence into an image with a fixed size of 16x8192 (512 patches) by default, with $p_h=p_w=16$.

\paragraph{Downstream Evaluation.} 
Our objective is to demonstrate that in-context length can be extended using visual tokens, thereby enhancing the understanding of complex multimodal documents. Consequently, we focus primarily on tasks related to long-context understanding.

To evaluate the long-context understanding ability, we adopt the few-shot evaluation setting in Flamingo~\cite{flamingo}. 
We report answer accuracy on the OK-VQA~\cite{okvqa}, TextVQA~\cite{textvqa}, VizWiz~\cite{vizwiz}, and VQAV2~\cite{vqav2}. 
Additionally, we assess performance on captioning tasks using  COCO~\cite{coco} and Flickr30K~\cite{flickr30k}.
Moreover, we also propose a setting named \textbf{text-only in context few-shots} to explore text-only in-context evaluation. 
For this setting, we use in-context sampling without visual input to generate long-context inputs and the visual input is not observed by the model. 

In order to illustrate the impact of having long in-context text, we  evaluate the model for document understanding on DocVQA~\cite{docvqa} and OCR VQA~\cite{ocrvqa}.
Lastly, we introduce a new task, sequential multimodal document retrieval. This dataset is based on the existing interleaved OBELICS~\cite{obelics} dataset. Further details are provided in the appendix.

\begin{table*}[t]
\centering
\small
\scalebox{.98}{
\begin{NiceTabularX}{\linewidth}{y{40}y{20}x{24}x{30}|x{18}x{18}x{18}x{18}x{18}x{18}x{18}x{18}}
\CodeBefore
\rowlistcolors{1}{background,background,background,background,background,background,background,background,softblue,softblue,softblue,background,background,background,softblue,softblue,softblue,background,background,background,background,background,background,softblue,softblue,softblue}
\Body
\toprule
    \bf Method & \bf Text  &\bf  ICL Tokens$\uparrow$&\bf Shots  & \multicolumn{4}{c}{\bf VQA} & \multicolumn{2}{c}{\bf Caption} & \bf Classi. & \bf Mean \\
    & & &  & okvqa & textvqa & vizwiz & vqav2 & coco & flickr &HM& \\
    \midrule
    \bf \multirow{3}{*}{\parbox[c]{2cm}{Open-Flamingo MOE~\cite{openflamingo}$\dag$}} & \multirow{3}{*}{\parbox[c]{1cm}{Raw Text}}  & \multirow{3}{*}{256}
    & 0&40.2&21.3&23.3& 47.8&82.3 &59.4 & 60.4 & 47.8 \\
    &&&4 &42.5&22.2&32.2&49.8&90.5&63.5&63.8 & 52.1\\
    &&&32 & \bf 46.8&23.2&40.5& 49.9 &\underline{98.2}&\underline{66.2}& \bf 66.0 &\underline{55.8} \\
    \hline
    \bf \multirow{3}{*}{\parbox[c]{1.5cm}{+ \ModelName}}   &  \multirow{3}{*}{\parbox[c]{1cm}{+ Rendered Image}}  & \multirow{3}{*}{\bf 2048}
    & 0& 39.5 & 26.4 & 26.3 & 48.5 & 84.4 & 60.5 & 62.2 & 49.7 \\
    &&& 4 & 44.3 & 28.9 & 32.0 & \underline{50.3} & 94.2 & 65.3 & \underline{65.5} & 54.4 \\
    &&& 32 & \underline{46.3} & \bf 31.2 & \bf 41.2 & \bf 51.0 & \bf 101.3 & \bf 68.4 & 65.2 & \bf 57.8 \\
    \bottomrule
\end{NiceTabularX}
}
    \caption{
    \small \textbf{Increasing in-context text length with \ModelName significantly improves performance on multi-modality downstream tasks.}
    The model is pre-trained with a 56B MOE model.
    ICL stands for in-context text length.
    HM is short for hatefulmemes.
    With \ModelName, we increase the ICL from 256 to 2048, leading to clear improvements over the baseline.
    $\dag$ indicates our implementation.
    }
    \vspace{-1em}
    \label{tab:flamingo_moe}
\end{table*}

\begin{table*}[t]
\centering
\small
\scalebox{.95}{
\begin{NiceTabularX}{\linewidth}{ly{20}x{24}x{30}|x{18}x{18}x{18}x{18}x{18}x{18}x{38}}
\CodeBefore
\rowlistcolors{1}{background,background,background,background,background,background,background,background,softblue,softblue,softblue,background,background,background,softblue,softblue,softblue,background,background,background,background,background,background,softblue,softblue,softblue}
\Body
\toprule
    \bf Method  & \bf Text Source & \bf Text Tokens&\bf T-Shots  & \multicolumn{4}{c}{\bf VQA} & \multicolumn{2}{c}{\bf Caption} & \bf Mean \\
    & & &  & okvqa & textvqa & vizwiz & vqav2 & coco & flickr & \\
    \midrule
    \bf \multirow{3}{*}{\parbox[c]{2cm}{Open-Flamingo9B Baseline~\cite{openflamingo}}$\dag$}  &  \multirow{3}{*}{\makecell{Raw \\ Text}}  & \textcolor{softlowcost}{10}
    & 0& 18.1 & 14.8 &21.5&26.5& 40.1 & 32.1 & 25.5 \\
    &&\textcolor{softmediumcost}{62}& 4 &\underline{23.8} & 18.1 & \underline{23.7} &\bf 40.5 & 57.5 & 35.3 & 33.2\greenp{7.7$\uparrow$} \\
    &&\textcolor{softhighcost}{426} & 32& \bf 25.2 & 16.4 & \bf 25.5 &34.6 & \bf 66.1 & \bf 38.5 & \bf 34.4\greenp{8.9$\uparrow$} \\
    \hline
    \bf \multirow{3}{*}{\parbox[c]{.5cm}{+\ModelName}}   &  \multirow{3}{*}{\parbox[c]{1cm}{Rendered Image}}  & \textcolor{softlowcost}{10}
    & 0& 16.2 & 16.8 & 15.4 & 30.6 & 42.3 & 33.5 & 25.8 \\
    &&\textcolor{softlowcost}{10}& 4 & 17.2 & \underline{21.8} & 19.7 & 35.2 & 52.4 & 35.2 & 30.3\greenp{4.5$\uparrow$}  \\
    &&\textcolor{softlowcost}{10}& 32 & 21.3 & \bf 22.6 & 21.5 & \underline{38.8}   & \underline{60.3} & \underline{37.0} & \bf \underline{33.6\greenp{7.8$\uparrow$}}\\
    \bottomrule
\end{NiceTabularX}
}
    \caption{ \small
    \textbf{\ModelName effectively incorporates in-context text with visual tokens, demonstrating significant performance improvements with consistent token usage}. 
    Here, T-shots refer to text-only in-context examples. Tokens indicate the length of the input to the LLM. Text source describes the preprocessing method for in-context examples. $\dag$ denotes our implementation on 180M pretraining data.
    }
    \label{tab:flamingo_icl_test}
\end{table*}

\begin{table*}[t]
\centering
\small
\begin{NiceTabularX}{\linewidth}{ll|ccc}
\CodeBefore
\rowlistcolors{1}{background,background,background,background,background,background,background,background,softblue,softblue,softblue,background,background,background,softblue,softblue,softblue,background,background,background,background,background,background,softblue,softblue,softblue}
\Body
  \toprule
    \bf Method  & \bf Text Source  & \multicolumn{2}{c}{\bf DocVQA} & \bf OCR VQA \\
    & &  val & test \\
    \midrule
    \bf \multirow{1}{*}{\parbox[c]{5cm}{Open-Flamingo-9B Baseline~\cite{openflamingo}}} &  \multirow{1}{*}{Raw Text} 
    & 45.3 & 48.2 & 51.5\\
    \midrule
    \bf \multirow{1}{*}{\parbox[c]{2cm}{+\ModelName}} & \multirow{1}{*}{Rendered Image} 
    &\bf 48.5\greenp{3.2$\uparrow$} & \bf 52.2\greenp{4.0$\uparrow$} & \bf 58.4\greenp{6.9$\uparrow$} \\
    \bottomrule
\end{NiceTabularX}
    \caption{
    \small
    \textbf{\ModelName clearly boosting the baseline on document understanding tasks.}
    }
    \label{tab:docvqa}
\end{table*}

\begin{figure}
    \centering
    \includegraphics[width=\linewidth]{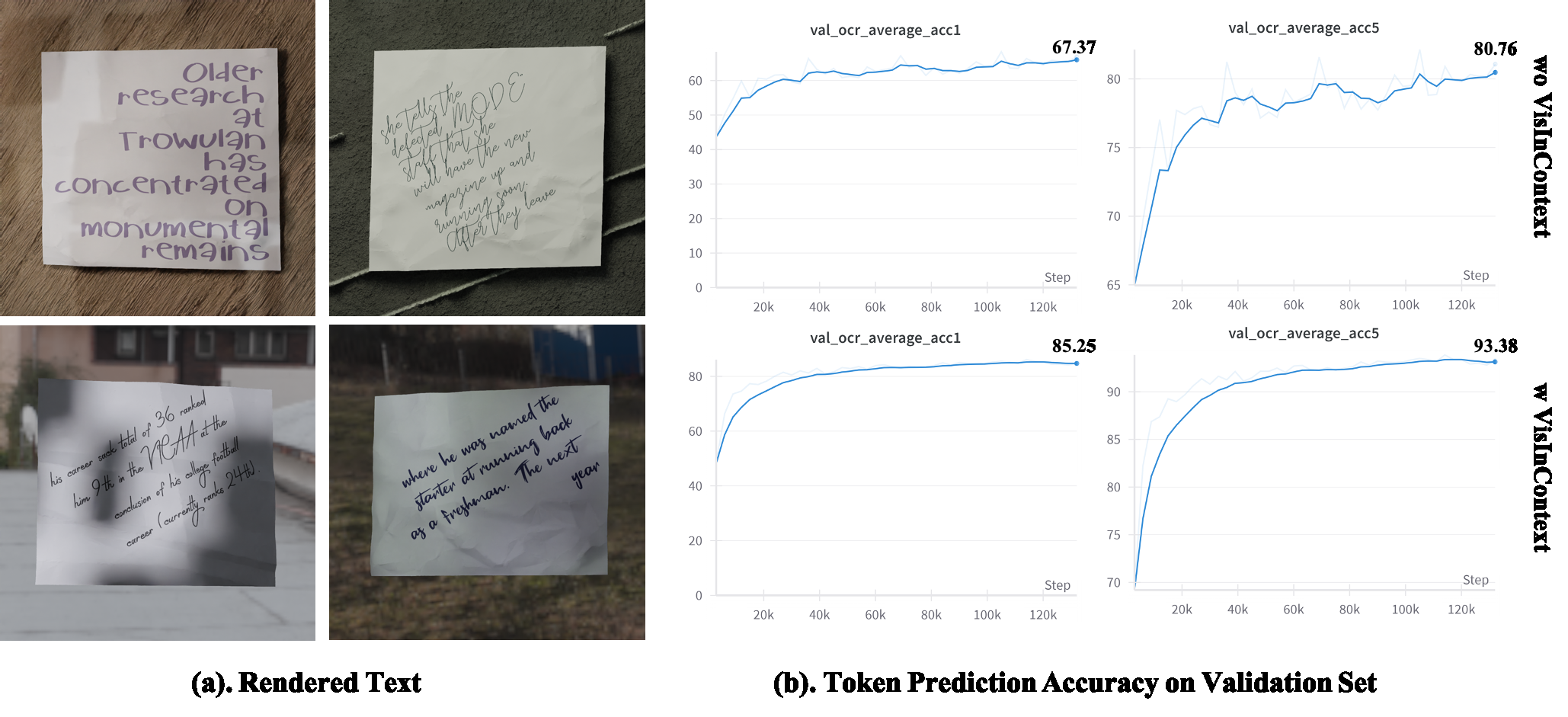}
    \vspace{-1em}
    \caption{
    \small
    \textbf{\ModelName significantly improves the OCR ability of LLM}.
    We present the Rendered Text~\cite{renderedtext2023} images and the corresponding next-word prediction accuracy on the validation set. 
    Using the same pre-training steps, \ModelName achieves significantly better results in predicting words in visual images, even when the fonts are difficult to recognize.
    }
    \vspace{-1em}
    \label{fig:rendered_text_validation}
\end{figure}

\subsection{In-context Few-shot Evaluation}
\paragraph{Impact of Extended In-Context Text Length.}
Interleaved document datasets typically contain long texts. 
For instance, the OBELICS~\cite{obelics} dataset has an average token length of 815 tokens per document. 
Due to GPU memory constraints, Flamingo-like models~\cite{flamingo,obelics,openflamingo} only sub-sample 256 tokens during pretraining, which leads to a significant loss of context information.
We compare the baseline model pre-trained with 256 tokens, against our method with an increasing \textit{In-context Text Length} to 2048 tokens.
Table~\ref{tab:flamingo_moe} shows a clear advantage of \ModelName. For example, on TextVQA, accuracy improves from 23.2\% to 31.2\% with 32-shot. 
Similarly, the average model performance across all datasets show an increase from 55.8\% to 57.8\%.
These findings demonstrate that \textbf{\ModelName~effectively increases the \textit{In-context Text Length} to improve multi-modality understanding}.

\paragraph{Few-shot Evaluation with Text-only In-context Examples.}
As downstream tasks often differ in format from pretraining data, several works~\cite{flamingo,openflamingo,obelics} have tested the few-shot abilities of models using in-context examples. 
For instance, in the VQA dataset, a few question-and-answer pairs are provided as in-context examples with visual signals. 
However, for zero-shot evaluation, two question-and-answer pairs are added as in-context examples without visual signals in ~\cite{flamingo,openflamingo,obelics}.
Follow the zero-shot setting, we examine the effect of having text-only in-context examples and extend it to multi-shot setting, by leaving out the corresponding images (See Appendix for more details). 
We compare model performance of the baseline Open-Flamingo 9B and our method under the same setting, where the differences lie in how these text-only in-context examples are processed. Specifically, Open-Flamingo directly takes in them as text tokens, while \ModelName takes in the corresponding rendered text images.

\setlength{\intextsep}{0pt} 
\setlength{\columnsep}{5pt} 

\begin{wrapfigure}[14]{l}{0.4\textwidth} 
    \centering
    \includegraphics[width=\linewidth]{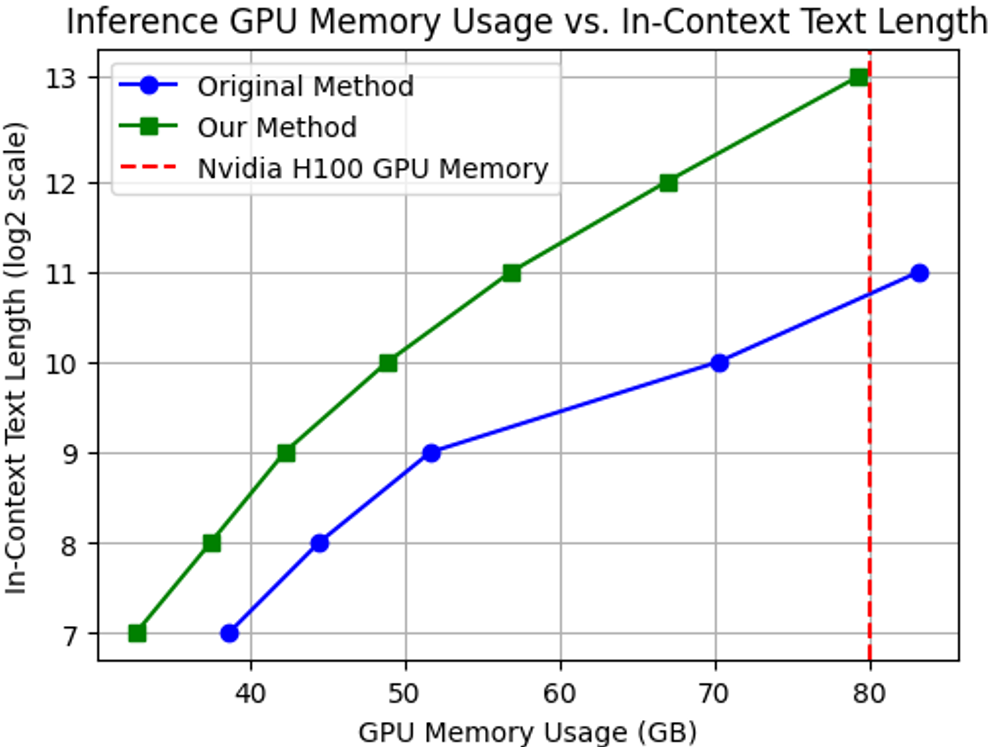}
    \caption{\small 
    \ModelName extends the in-context text length of MOE based MLLM from 1k to 9k at inference stage.
    }
    \label{fig:inference_cost}
\end{wrapfigure}

Table~\ref{tab:flamingo_icl_test} summarizes the results across four VQA benchmarks and two captioning benchmarks. 
Notably, compared to the text-only 0-shot setting, our \ModelName with 32-shot significantly is improved on  all VQA and captioning benchmarks considered. 
Though the 32-shot performance of \ModelName is slightly lower than the competing baseline, we cut down the input tokens to the LLM from 426 to only 10 \textit{Text Token Length}, which lead to significant reduction in the inference cost.
These outcomes highlight two key points: \emph{i}. \textbf{\ModelName can effectively understand text rendered in images}.
\emph{ii}. Text rendered as images can be comparably effective as raw text, when used as text-only in-context examples.

\paragraph{Comparison on Inference Cost.}
We then analyze the inference cost of \ModelName and compare to the baseline. 
Both models are based on a 56B MOE LLM with a batch size of one to explore the maximum manageable \textit{In-context Text Length}. 
The results, shown in Figure~\ref{fig:inference_cost}, demonstrate that the \textit{In-context Text Length} can be extended up to 9192 tokens for the 56B MOE model on 80GB H100 GPUs with our method at inference stage. 
This result highlights the efficiency and advantages of \ModelName, also show its potential in understanding very long document.

\subsection{Document understanding}
In this section, we evaluate the model on document understanding tasks. Unlike common vision-language tasks that usually short-form pairs, this task requires comprehension of long and complex document data. 
We evaluate our model on DocVQA and OCRVQA. All document images are of size $384 \times 384$.
Following Pix2Struct~\cite{lee2023pix2struct}, we finetune the model on DocVQA train data and report performance on the average normalized Levenshtein similarity (ANLS) metric.

Results in Table~\ref{tab:docvqa} show that our method significantly outperforms the baseline. For instance, we achieve a 6.9\% improvement on OCRVQA.
To further analyze why our method enhances document understanding, we present the validation accuracy of the LLM on the Rendered Text~\cite{renderedtext2023} dataset during pretraining in Figure~\ref{fig:rendered_text_validation}. We observe a substantial improvement in next word prediction accuracy, with top-1 accuracy increasing from 67.37\% to 85.25\% (a 16\% improvement) and top-5 accuracy rising from 80.76\% to 93.38\%.
These findings indicate that the \textbf{ LLM can effectively understand text embedded in visual signals with \ModelName}.

\begin{figure}[t]
    \centering
\includegraphics[width=\linewidth]{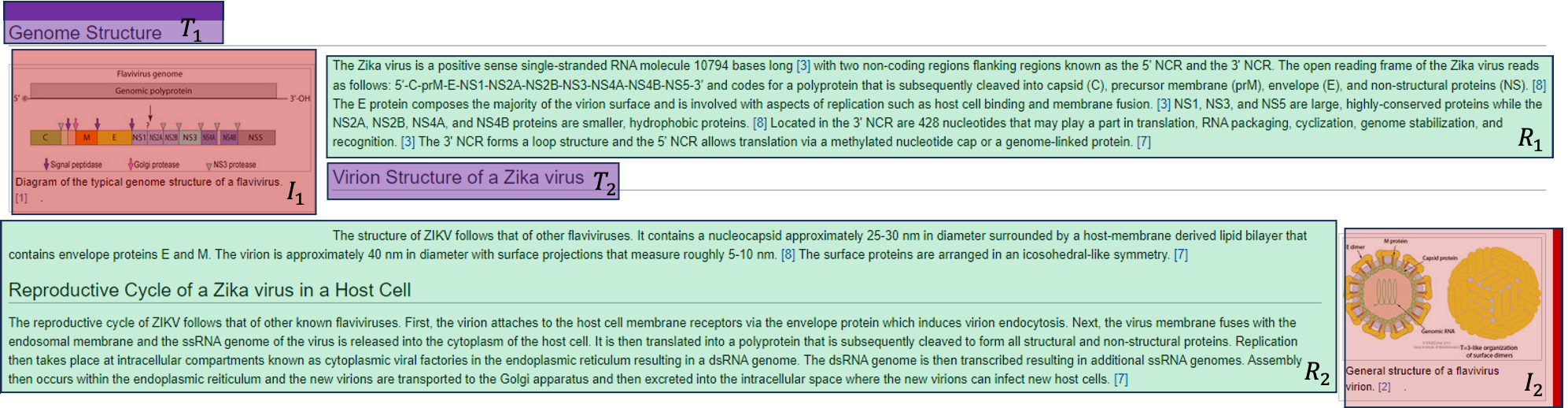}
    \caption{
    \small
    \textbf{Sequential multi-modal retrieval example.}
    The input sequence is $I_1,T_1,R_1,I_2,T_2,R_2$ that from interleaved document in OBELICS~\cite{obelics} dataset.}    
    \label{fig:rendered_text}
\end{figure}
\begin{table}[t]
    \centering
    \small
    \begin{tabular}{ccc|cc}
    \toprule
      \bf Visual Input &  \bf Text Input &\bf Surrounding Text Input  & \bf Seq-I & \bf Seq-T  \\
      \midrule
        Raw Image & Raw Text &-& 16.3 & 64.8\\
        Raw Image & Raw Text &Raw Text& 18.9 & \bf 67.5\\
        Raw Image & Raw Text&Rendered Text Image& \bf 22.7 & 66.5 \\
        \bottomrule
    \end{tabular}
    \caption{
    \small 
    \textbf{The model pretrain with \ModelName significantly improves sequence understanding ability.
    }
    We report the sequence retrieval result on OBELICS-Hybrid6.
    }
    \label{tab:seq_retrieval}
\end{table}


\begin{table}[t]
\centering
\scalebox{.95}{
\begin{tabular}{cc|ccc}
\toprule
    \bf Method  & \bf Pretrain Text Source   & \multicolumn{1}{c}{\bf Task}  \\
    & & DocVQA-val \\
    \midrule
    \bf \multirow{1}{*}{\parbox[c]{3cm}{FuYu9B~\cite{adept_fuyu8b}$\dag$}}  &  \multirow{1}{*}{Raw-Text} &   42.3  \\
    \hline
    \bf \multirow{1}{*}{\parbox[c]{3cm}{+ \ModelName}}   & \multirow{1}{*}{+Rendered Image}  & \bf 44.5 \greenp{2.2$\uparrow$}\\
    \bottomrule
\end{tabular}
}
    \caption{
    \textbf{Pretraining with \ModelName helps on long-context understanding task for FuYu model}.
     $\dag$ means our implementation on 180M data.
    }
    \label{tab:fuyu_vqa}
\end{table}

\subsection{Sequential Multi-modal Retrieval}

In order to further analyze the benefit of having long text context in multimodal modeling, we propose a new task -- Sequential Multimodal Retrieval (SMR), based on document data from interleaved OBELICS~\cite{obelics} dataset. The document is composed of interleaved data, consisting of images and texts arranged in a meaningful sequence.

We show one sample in Figure~\ref{fig:rendered_text} and define the input and output of this task as below: \textbf{Input:}  Given a pair of content items, an image and a corresponding text $(I_1, T_1, R_1, I_2, T_2, R_2)$, from a document $D$. $I$ is Image, $T$ is the matched text and $R$ is the surrounding text.
\textbf{Output:} 
The task is to retrieve the next image $I_{2}$ and the next text $ T_{2}$ in the sequence. 
Named as Seq-I and Seq-T, correspondingly.

We sample the first 1K documents that contain data like $I_1,T_1,R_1,I_2,T_2,R_2$ from OBELICS~\cite{obelics} and named it as OBELICS-Hybrid6, which have at least three frames and three texts.
(See appendix for more details.)
This task encourages the model to leverage the contextual and semantic relationship in interleaved sequences to effectively predict and retrieve the subsequent pair.

To enable our model with retrieval,  we follow CosMo~\cite{cosmo} to add a simple contrastive head between visual embedding and language embedding from the middle layers. Recall that visual embeddings are either from raw images or rendered images or the addition of the two in our method.
Table~\ref{tab:seq_retrieval} reports the results from our model with several input variants.
We observe taking surrounding text input as rendered text image performs much better on the Sequence to image retrieval, while on par on Sequence to text retrieval, when compared with taking surrounding text input as raw text. These results further support the designs of \ModelName~ in the context of document understanding.

\subsection{Extension to MLLM with Linear Embedding}

Beyond utilizing the visual encoder, some works~\cite{vilt,adept_fuyu8b} also employ linear embedding to extract visual features directly from raw images.
To show the generality of our method, we also explore FuYu~\cite{adept_fuyu8b} model as a baseline and integrate \ModelName into the model. (See the appendix for more details.)
As indicated in Table~\ref{tab:fuyu_vqa}, our method is successful in improving the performances on DocVQA dataset that require long-context understanding. 
\vspace{1em}

\begin{table}[h]
    \centering
\small
\scalebox{.95}{
    \begin{tabular}{ccc|cccc}
    \toprule
       Text Image & Token Masking & TCCL & Ok-VQA & TextVqa & VizWiz & VqaV2  \\
       \midrule
        & && 11.5 & 15.3 & 8.7 & 24.2 \\
        \checkmark && &11.3 & 15.0 & 9.4 & 30.1  \\
        \checkmark  & \checkmark  &&\bf 17.8&18.3& 15.3 & 33.5\\
         \checkmark  &   &\checkmark& 13.5&15.3& 10.3 & 30.9\\
        \checkmark & \checkmark& \checkmark   &  17.2& \bf 21.8 & \bf 19.7 & \bf 35.2 \\
        \bottomrule
    \end{tabular}
    }
    \caption{\small Ablation study of the component in our pipeline for text-only 4-shot example.}
    \label{tab:ablation}
\end{table}

\begin{table}[h]
    \centering
    \footnotesize
    \begin{minipage}{0.45\textwidth}
        \centering
        \scalebox{.9}{
        \begin{tabular}{c|ccccc}
        \toprule
          \bf Font Size  &\bf 4 &\bf 6 &\bf 8 &\bf 10 &\bf 12  \\
          \midrule
           TextVQA & 15.4 & 17.2  & 18.5 &\bf 21.8 & 20.3 \\
          DocVQA & 39.8 &42.5 &\bf 45.6&44.3&36.2\\
             \bottomrule
        \end{tabular}
    }
        \caption{
        \small
        \textbf{Font size ablation.}
        We report the result on DocVQA val dataset.
        }
        \label{tab:font_size_abl}
    \end{minipage}%
    \hfill
    \begin{minipage}{0.45\textwidth}
        \centering
        \scalebox{.9}{
        \begin{tabular}{c|ccccc}
        \toprule
          \bf Dataset  &\bf 2 &\bf 4 &\bf 8 &\bf 16 &\bf 32  \\
          \midrule
            TextVQA & \bf 21.8 & 20.5 &21.3 &18.5 &15.3 \\
            DocVQA & \bf 44.3&43.2&39.4&40.5&36.6\\
             \bottomrule
        \end{tabular}
        }
        \caption{
        \small
        \textbf{
            Font interval thresh ablation.
        }
        Larger thresh leads to few texts in general.
        }
        \label{tab:font_interval_abl}
    \end{minipage}
\end{table}

\subsection{Ablation Study}
\paragraph{Ablations on Model Design.} We conduct ablation studies on the following modeling components to demonstrate their effectiveness: Text Image, TCCL, and Token Masking. 
Results are detailed in Table~\ref{tab:ablation}, which reveal two findings: 
1. Token Masking is crucial for the model to learn from rendered text images. 
Without Token Masking, the model can only perform comparably to the baseline.  
Forcing the model to learn text semantics from rendered text images via token masking significantly improves model performance.
2. Utilizing TCCL with Token Masking yields better performance than using Token Masking alone.

\paragraph{Ablations on Font Size and Interval Threshold.} As shown in Table~\ref{tab:font_size_abl}, optimal performance varies with changes in font size. 
We found that adjusting the font size impacts performance similarly to altering the patch size—both methods effectively increase the contextual information within each patch. 
We prefer modifying the font size over the patch size because it allows for more intuitive adjustments. Our findings indicate that the model does not need a highly detailed understanding of each word to perform effectively.

Another important factor is the font interval threshold. 
As shown in Table~\ref{tab:font_interval_abl}, we observed that a too-large interval leads to inferior results. 
This is intuitive because a larger threshold results in fewer texts in the rendered text image.
\section{Related Work}

\paragraph{Multimodal Language Models.}
Current mainstream Multimodal Large Language Models (MLLMs)~\cite{blip2,palme,clip,coca,llava,qwen} leverage the capabilities of Large Language Models (LLMs)~\cite{gpt3,llama2} due to their strong reasoning abilities, as demonstrated by recent advancements. 
These models typically adopt one of two primary designs for integrating visual information. The first approach involves the effective adaptation of visual representations, which are acquired via a separate visual encoder, into the text-based LLM framework like CLIP, GIT, and BLIP2~\cite{clip,git,blip2}.
The representative method in this category incorporates visual representations into the language model using cross-attention, as seen in the Flamingo series models~\cite{flamingo,openflamingo,obelics}.
Along this line, recently some works like LLaVA~\cite{llava}, 
EMU2~\cite{emu}, 
InternVL~\cite{internvl}, DeepSeeker~\cite{deepseek}, and QWen~\cite{qwen} lead to superior results on multi-modality tasks with supervised finetuning on high-quality data. The second approach uses visual embeddings directly as input "tokens" for the LLMs, bypassing the traditional use of a separate visual encoder. This method processes visual patches with a linear layer and uses the resulting embeddings as direct inputs to the LLM, as implemented in models like ViLT~\cite{vilt} and FuYu~\cite{adept_fuyu8b}. 
This strategy omits the need for an additional visual encoder and simplifies the architecture.

In this work, we adopt the Flamingo~\cite{flamingo} architecture as our main baseline for the following reasons: First, the Flamingo model emphasizes in-context few-shot learning ability and designs comprehensive few-shot evaluation strategies. Second, our focus is on extending the in-context text length during pre-training rather than on supervised fine-tuning.

\paragraph{Enhancing Text Understanding through Visual Inputs.}
Traditional text tokenization processes raw text efficiently, but it faces challenges such as vulnerability to spelling errors and limited cross-lingual transferability~\cite{rust2023language,gao2024improving}. 
These issues have prompted the exploration of tokenizer-free models, which aim to improve robustness and facilitate better cross-language applicability. For instance, a single spelling error can lead to entirely different tokens using traditional tokenization methods, impacting model performance.

Recent developments have seen innovative approaches like the Pixel model~\cite{rust2023language}, which proposes processing text as an image using both an image encoder and an image decoder. This approach has sparked a series of studies that process not only textual data but also images, charts, and tables through a unified visual input system~\cite{lee2023pix2struct,rust2023language,clippo,gao2024improving}. These models are trained on a diverse array of visual data, such as webpage screenshots and user interface images, sourced extensively from the internet. They are specifically designed to handle visually-situated text in an end-to-end manner, offering the potential to support a wide range of applications.

\paragraph{Long Context Modeling.}
The challenge of incorporating more tokens into LLMs is an active area of research~\cite{child2019generating,beltagy2020longformer}. 
Common approaches involve novel self-attention mechanisms~\cite{ring} or memory banks~\cite{memorizingtransformer}. 
Some works~\cite{korthikanti2023reducing} exploit tensor parallelism or sequence parallelism to reduce memory costs. In multi-modality research, closed-source models like Gemini~\cite{gemini} and GPT-4V~\cite{openai_gptv_system_card} support long context inference up to millions of tokens. Open-source models such as MA-LMM for Long-Term Video Understanding~\cite{malmm} can process up to one hour of video using a long memory bank. 
The most relevant work Large World Model~\cite{lvm} extends token length using Ring Attention.

In contrast to these methods, our method utilizes off-the-shelf LLMs and compresses text tokens into visual tokens for efficient processing. 
Our method is complementary to these existing techniques and can be integrated with them to achieve lower computational cost and longer context length.

\section{Conclusion and Limitations}
This paper centers on multi-modality learning and addresses the in-context length limitations presented by heavy computational cost of LLMs in MLLMs. 
Our contribution is a novel and efficient method named \ModelName, which enables the model to perceive long text context as rendered text images. Comprehensive experiments show that \ModelName is effective on conventional in-context few-shot evaluations and document 
understanding, while being much more efficient.

One limitation of our method is, currently our method requires processing a fixed size image even for brief texts. 
In future work, we plan to dynamically reduce token counts with variable image sizes by retaining only non-empty tokens during pre-training. 
We aim to expand this method to additional tasks and encourage the community to further explore this direction.

{\small
\bibliographystyle{unsrt}
\bibliography{reference}
}

\end{document}